# A Multi-modal Fusion Framework Based on Multi-task Correlation Learning for Cancer Prognosis Prediction⋆

Kaiwen Tan, Weixian Huang, Xiaofeng Liu, Jinlong Hu and Shoubin Dong*

*Communication and Computer Network Lab of Guangdong, School of Computer Science and Engineering, South China University of Technology, Guangzhou, China*

## ARTICLE INFO

## ABSTRACT

Morphological attributes from histopathological images and molecular profiles from genomic data are important information to drive diagnosis, prognosis, and therapy of cancers. By integrating these heterogeneous but complementary data, many multi-modal methods are proposed to study the complex mechanisms of cancers, and most of them achieve comparable or better results from previous single-modal methods. However, these multi-modal methods are restricted to a single task (e.g., survival analysis or grade classification), and thus neglect the correlation between different tasks. In this study, we present a multi-modal fusion framework based on multi-task correlation learning (MultiCoFusion) for survival analysis and cancer grade classification, which combines the power of multiple modalities and multiple tasks. Specifically, a pre-trained ResNet-152 and a sparse graph convolutional network (SGCN) are used to learn the representations of histopathological images and mRNA expression data respectively. Then these representations are fused by a fully connected neural network (FCNN), which is also a multi-task shared network. Finally, the results of survival analysis and cancer grade classification output simultaneously. The framework is trained by an alternate scheme. We systematically evaluate our framework using glioma datasets from The Cancer Genome Atlas (TCGA). Results demonstrate that MultiCoFusion learns better representations than traditional feature extraction methods. With the help of multi-task alternating learning, even simple multi-modal concatenation can achieve better performance than other deep learning and traditional methods. Multi-task learning can improve the performance of multiple tasks not just one of them, and it is effective in both single-modal and multi-modal data.

## 1. Introduction

Morphological attributes from histopathological images and molecular profiles from genomic data are important information to drive diagnosis, prognosis, and therapy of cancers. Traditionally, pathologists examine and evaluate patterns, textures, and morphology of cells and tissues in histopathological images to formulate a diagnosis or prognosis. With recent advancements in medical imaging and computing, many histopathological image analysis algorithms [32] have been proposed for survival analysis [35, 43] and grade classification [22, 40, 26]. In comparison with human inspection, these computing methods are more objective, and have great potential to improve accuracy and efficiency. Besides histopathological images, genomic data such as mRNA expression profiles and somatic mutation are also widely used for predicting the prognosis of cancers [45, 17, 38].

Histopathological images and genomic data are heterogeneous and high-dimensional data. A typical pathological whole slide image (WSI) contains $100,000 \times 100,000$ pixels, and genomic data such as mRNA expression data contains more than ten thousands feature (i.e., genes). But meanwhile, they capture the patient's characteristics at different scales and from different perspectives and are natu-rally considered to contain complementary information for cancer prognosis prediction [31]. For example, according to the World Health Organization (WHO) classification of central nervous system [20], diffuse gliomas are first classified into one of three different subtypes (i.e., IDH wild-type astrocytoma, IDH mutant astrocytoma, and oligodendroglioma) based on the IDH mutation and 1p19q co-deletion status. These subtypes are then assigned WHO grade II-IV based on the histopathological images, which indicate different degrees of malignancy.

To provide a more objective, accurate and efficient result, many multi-modal fusion methods have been proposed to address above issues (i.e., heterogeneity and high-dimensionality), and study the complex mechanisms of cancers. Most of them achieve comparable or better results from previous single-modal methods. We divide these methods into two stages: single-modal feature extraction/representation learning and multi-modal fusion.

For single-modal feature extraction/representation learning, its goal is to extract important features or feature's representations closely related to the target and reduce the feature dimension. It's a critical step to facilitate subsequent fusion. Some methods utilize unsupervised learning, statistics or bioinformatics tools to extract the features from histopathological images and genomic data. For example, Cheng *et al.* [5] and Shao *et al.* [30] first use an unsupervised segmentation method to segment nucleus, and then they extract cell-level features for each segmented nucleus. Finally, they aggregate these cell-level features extracted from a patient into patient-level features. Meanwhile, they carry out gene co-

---

⋆This work was supported by National Natural Science Foundation of China (61976239) and Zhongshan Innovation Foundation of High-end Scientific Research Institutions (2019AG031).

*Corresponding author

kwtan93@mail.scut.edu.cn (K. Tan); sbdong@scut.edu.cn (S. Dong)

ORCID(s): 0000-0002-2770-4863 (K. Tan)

expression network analysis (GCNA) to cluster genes into co-expressed modules and summarize each module as an eigengene. Deep learning is another widely used technology to extract features, which is usually called representation learning. For example, Hao *et al.* [12] and Ning *et al.* [23] use convolutional neural networks (CNNs) to extract representations of histopathological images. Mobadersany *et al.* [21] propose a CNNs-based method called survival convolutional neural networks to extract representations of histopathological image related to survival outcome. Pathomic Fusion [4] extracts representations of histopathological image by a 19-layer Visual Geometry Group (VGG-19), which is based on CNNs and can be fine-tuned using preexisting weights trained on ImageNet. Meanwhile, Pathomic Fusion uses a Feedforward Self-Normalizing Network [14] called Genomic SNN (GSNN) to learn a low-dimensional feature-rich representation for genomic data. However, VGG-19 has 143 million parameters and 19.6 billion FLOPs, it requires a huge space to store the network and a long time to train and test. Therefore, in our model, we use ResNet-152 instead of VGG-19 to learn the representation of histopathological images. Because ResNet-152 has fewer parameters (60 million v.s. 143 million), lower complexity (11.3 billion FLOPs v.s. 19.6 billion FLOPs) but more deeper than VGG-19 (152 layers v.s. 19 layers) [13]. For genomic data (e.g., mRNA expression data), traditional methods such as fully connected neural network (FCNN) and GSNN ignore the interaction between features (e.g., genes), so we use sparse graph convolutional network (SGCN) [36] to integrate genegene interactions to learn the represention of mRNA expression data.

For multi-modal fusion, Multiple Kernel Learning (MKL) has been widely used [37, 33, 47], which combines several kernels into one optimized kernel with arithmetic mean or geometric mean for prognosis prediction. Canonical Correlation Analysis (CCA) can capture the intrinsic relationship among multiple modalities. Shao *et al.* [30] introduce survival information into CCA to enhance the power of survival analysis. Non-negative matrix factorization (NMF) is an effective dimensionality reduction method. Deng *et al.* [8] propose a multi-constrained joint NMF (MCJNMF) model to study the relationship between FEG-PET image and DNA methylation data in a common feature space. They apply the network regularization as prior knowledge, which can improve the accuracy of factorization. Recently, Deng et al. extend their MCJNMF to more modalities, which is called multi-dimensional constrained joint nonnegative matrix factorization (MDJNMF) [9]. Based on the morphologic features extract from histopathological images and eigengenes extract from mRNA expression profiles, Cheng *et al.* [5] build a LASSO-regularized Cox proportional hazards model to predict the risk of each patient. Different from parameterdependent methods such as Log-rank test and LASSO, Ning *et al.* [23] utilize a parameter-free multivariate feature selection method called block filtering post-pruning search (BFPS) [25] to fuse the features from multiple modalities for survival analysis. Recently, Ning *et al.*[24] propose a relation-induced multi-modal shared representation learning (RIMM-SRL) for Alzheimer's disease diagnosis, which utilized regularization and relation-induced constriction. With recent advancements in deep learning, various deep learning-based methods have been proposed to fuse multi-modal data. For example, Mobadersany *et al.* [21] combine histopathological image features and genomic variables into a vector by concatenating, then the vector is fed into fully connected layers and Cox model for survival analysis. Similarly, Hao *et al.* [12] concatenate high-level representations of the histopathological and genomic data into a Cox proportional hazards regression model for survival analysis. Venugopalan *et al.* [39] propose Deep Model that used stacked denoising autoencoders to extract features from clinical and genetic data, and used 3D-convolutional neural networks to extract features from imaging data, then concatenate these features for Alzheimer's disease diagnosis. By constraining the similarity of representations that learn from different modalities, Cheerla *et al.* [3] develop an unsupervised encoder to compress multiple modalities into a single feature vector for each patient, which can be used in subsequent tasks such as survival analysis. Pathomic Fusion [4] utilizes Kronecker product to model inter-modality interactions and control the expressiveness of each modal via a gating-based attention mechanism for survival analysis or grade classification. However, these multi-modal methods have been restricted to a single task (i.e., survival analysis or grade classification), and thus neglect the correlation between different tasks.

Multi-task learning is a broadly used learning paradigm [42, 34, 46]. The aim of multi-task learning is to optimize several learning tasks simultaneously and leverage their shared information to help improve generalization and prediction of the model for each task. In the biomedical field, Qi *et al.* [28] utilize semi-supervised multi-task learning to predict Protein-Protein Interactions. Du *et al.* [10] propose a temporal multi-task sparse canonical correlation analysis method, which can identify time-consistent and time-dependent phenotypic and genotypic markers simultaneously. Xie *et al.* [44] present a domain-adversarial multi-task method for joint learning of the representations of heterogeneous data to predict novel therapeutic property of compounds. But there is no method considering multi-task learning for prognosis tasks, and most of the above methods are trained by a joint training scheme, which may not be suitable for prognostic multi-task learning.

In this paper, we present a multi-modal fusion framework (MultiCoFusion) based on multi-task correlation learning to combine the power of multiple modalities (i.e., histopathological images and mRNA expression data) and multiple tasks (i.e., survival analysis and grade classification). The architecture of MultiCoFusion is shown in Figure.1. The representations of histopathological images and mRNA expression data are first learned by a ResNet-152 and a sparse graph convolutional network (SGCN) respectively. The ResNet-152 is fine-tuned using pre-existing weights trained on ImageNet [29]. The SGCN can model the relations between fea-tures (i.e., genes) to improve the quality of representations of mRNA expression data. These representations are then fused by a fully connected neural network, which is also a shared feature extractor network for multi-task learning. Finally, our framework outputs the results of two tasks simultaneously. The framework is training by an alternate scheme. For each batch, only a loss function of a specific task is optimized. We systematically evaluate our framework using a glioma dataset, which is consist of Lower-Grade Glioma (LGG) and Glioblastoma (GBM) from The Cancer Genome Atlas (TCGA). Experiment results demonstrate that multi-modal and multi-task are both effective methods to improve model performance. The combination of multi-modal and multi-task can further improve the performance of multiple tasks simultaneously. The major contributions of this study are summarized as follows:

- We propose that not only multiple modalities, but also multiple tasks can improve the performance of cancer diagnosis and prognosis prediction. Based on this idea, we present a multi-modal fusion framework (MultiCoFusion) based on multi-task correlation learning to combine the power of multiple modalities and multiple tasks.

- We use sparse graph convolutional network (SGCN) to learn the representations of mRNA expression data via gene-gene interactions.

- We construct systematic experiments to evaluate the performance of our framework. Experiment results demonstrate that multi-task learning can improve the performance of multiple tasks simultaneously, and it is effective in both single-modal and multi-modal data.

## 2. Materials and Methods

### 2.1. Dataset

The glioma dataset contains histopatholgcial images, genomic and clinical data from TCGA-LGG and TCGA-GBM projects. We download the image and clinical data from Chen et al. [4], which contains 954 region-of-interests (ROIs) from histopathological whole slide images (WSIs) for 470 patients, and their ground-truth survival and histologic grade labels. These ROIs are curated by Mobadersany et al. [21] with 1,024×1,024 pixels. We download the mRNA expression data of Brain Lower Grade Glioma (TCGA, PanCancer Atlas) and Glioblastoma Multiforme (TCGA, PanCancer Atlas) from cBioPortal [2, 11]. Among the 470 patients, one patient (TCGA-06-0152) is missing mRNA expression data, so we remove it. The data used in our experiments contain 953 samples for 469 patients. In this dataset, there are three types of labels for cancer grade classification, i.e., Grade II, Grade III, and Grade IV, and they have 393, 408 and 152 samples respectively. In addition, a gene-gene interaction graph is constructed from HINT [7], and an adjacency matrix is calculated based on the graph with a self-loop. Specifically, the high-throughput and binary dataset of human is downloaded from HINT, which contains 62,435 PPIs. A gene interaction graph is constructed based on these PPIs, where nodes denote gene symbols and edges denote whether the connected nodes are interacting. Then the graph is converted to an adjacency matrix with self-loop. Because the mRNA expression features (i.e., genes) and the gene-gene interaction features are not exactly the same, we take the intersection of their features as final features. As a result, there are 10673 genes in the mRNA expression data, and the size of the adjacency matrix of gene-gene interactions graph is 10673×10673. All experiments in this paper are conducted using the same train-test splits as Mobadersany et al. [21] and Chen et al. [4], which are generated randomly with 80% training samples and 20% testing samples. We repeat the randomized assignment of train-test splits 15 times and use each of these train-test splits to train and evaluate all models.

### 2.2. Single-modal representation learning
#### 2.2.1. Representation learning of histopathologcial image

In this study, we use a CNNs-based method called ResNet-152 to extract deep features from 1,024×1,024 ROIs at 20× magnification. Each 1,024×1,024 ROI is randomly transformed to a 512×512 crop. Our model is trained on these crops. The key idea of ResNet-152 is to explicitly approximate a residual function via shortcut connections. Formally, the building block (called bottleneck) of ResNet-152 is defined as:

$$h_{l+1}^{(1)} = \sigma(F_1(h_l^{(1)}, \theta_l^{(1)}) + F_2(h_l^{(1)})) \tag{1}$$

where $h_l^{(1)}$ and $h_{l+1}^{(1)}$ denote input and output of $l^{th}$ bottleneck. The $\theta_l^{(1)}$ is a trainable parameter. $F_1$ is a residual function that consists of three 2D-convolutions with batch normalization and $F_2$ is a linear projection or an identity mapping. The $\sigma$ is an ReLU activation function. By stacking the bottleneck, a 152-layer residual network called ResNet-152 is obtained. We use the parameters pre-trained on ImageNet to initialize our model. The final output of ResNet-152 is represented as $Z^{(1)} \in \mathbb{R}^{(n \times 1000)}$, where $n$ is the number of samples and 1000 is the dimension of sample's representation.

#### 2.2.2. Representation learning of mRNA expression data

To learn a low-dimensional feature-rich representation of mRNA expression data, traditional methods such as Genomic SNN [4] and FCNN ignore the interactions between genes. Motivated by graph-embedded deep feedforward network [15], we use a sparsely connected graph convolutional network layer to model the interactions between genes via aggregating neighbourhood information in a gene-gene interaction graph, which is called sparse graph convolutional network (SGCN) [36] and defined as:

$$H^{(2)} = \sigma \left( X^{(2)} \left( A \odot W_1^{(2)} \right) \right) \tag{2}$$

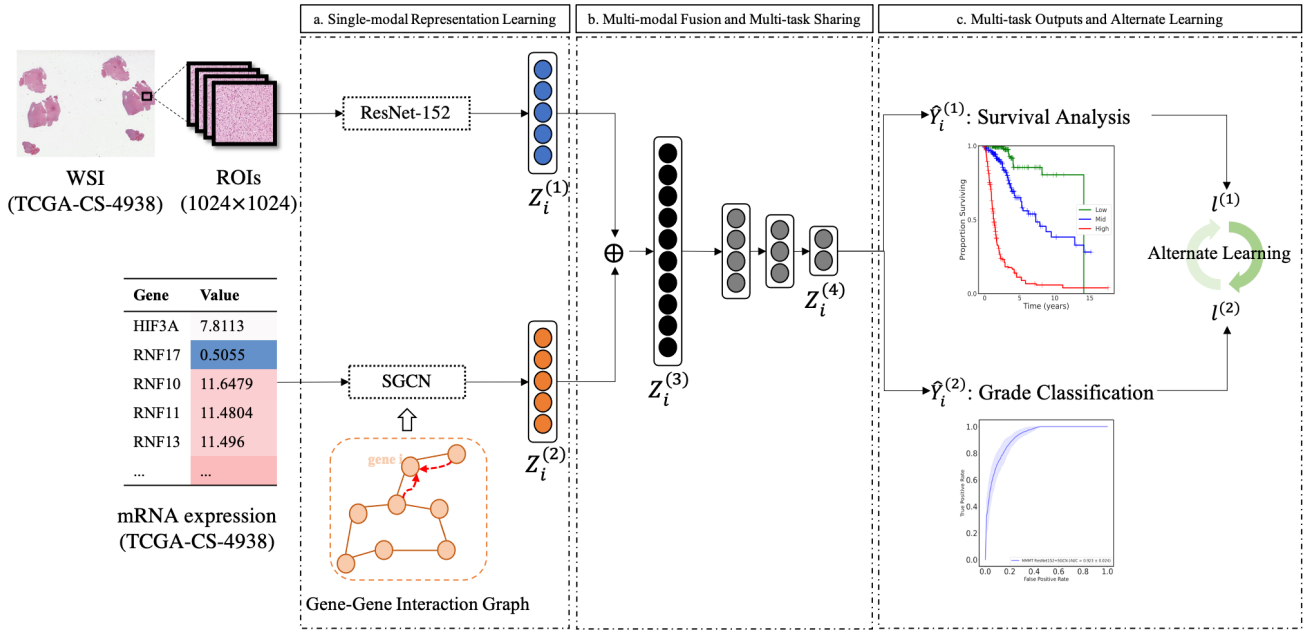

**Figure 1:** Illustration of MultiCoFusion framework. The histopathological whole slide images (WSIs) and mRNA expression data belong to patient TCGA-CS-4938. (a) Single-modal representation learning. A pre-trained ResNet-152 is used to learn the representations of ROIs. A sparse graph convolutional network (SGCN) is used to learn the representations of mRNA expression data. (b) Multi-modal fusion and multi-task sharing. First, the representations of ROIs and the mRNA expression data are concatenated, and then fuse through a three-layer fully connected neural network (FCNN), which is also a multi-task shared network. (c) Multi-task outputs and alternate learning. The results of multiple tasks output simultaneously. In training process, only a task's loss function is minimized at a time.

$$Z^{(2)} = \sigma\left(H^{(2)}W_2^{(2)}\right) \tag{3}$$

where $X^{(2)} \in \mathbb{R}^{(n \times p)}$ denotes the mRNA expression data, $n$ is the number of samples and $p$ is the dimension of each sample. $Z^{(2)} \in \mathbb{R}^{(n \times 1000)}$ is the representation of mRNA expression data. $A \in \mathbb{R}^{(p \times p)}$ is the adjacency matrix of the gene-gene interaction graph. $W_1^{(2)}$ and $W_2^{(2)}$ are trainable parameters. The operation $\odot$ is an element-wise multiplication operation called the Hadamard product. The Hadamard product between $A$ and $W_1^{(2)}$ produces a new weight matrix where only interacting genes have weights. Thus, the connection between the input layer and the first hidden layer is sparse, and the new matrix contains the structural information of the gene graph.

## 2.3. Multi-modal fusion and multi-task correlation learning

To fuse the representations (i.e., $Z^{(1)}$ and $Z^{(2)}$) learned from histopatholocgical image and mRNA expression data, we first concatenate them into a matrix $Z^{(3)} \in \mathbb{R}^{(n \times 2000)}$. Then a three-layer FCNN is employed to learn a fused representation, and meanwhile, the shared information between different tasks is learned by this network:

$$Z^{(4)} = G\left(Z^{(3)}, W_3^{(3)}\right) \tag{4}$$

where $W_3^{(3)}$ is a trainable parameter, and $G$ denotes the three-layer FCNN. $Z^{(4)} \in \mathbb{R}^{(n \times 32)}$ is the fused and shared repre-sentation. Based on this representation, a regressor and a classifier are used to output the results of survival analysis and grade classification respectively. The regressor is a two-layer FCNN following a sigmoid function. The classifier is a two-layer FCNN following a log-softmax function. The final results of the regressor and classifier are represented as $\hat{Y}^{(1)} \in \mathbb{R}^{(n \times 1)}$ and $\hat{Y}^{(2)} \in \mathbb{R}^{(n \times k)}$ respectively, where $k$ is the number of categories for histologic grade.

## 2.4. Alternate Training

The loss function of survival analysis is Cox partial likelihood [41]:

$$l^{(1)} = \sum_{i=1}^{n} \delta_i \left( \hat{Y}_i^{(1)} - log\left( \sum_{j \in \Omega(t_i)} \exp(\hat{Y}_j^{(1)}) \right) \right) \tag{5}$$

where $\delta_i$ is an indicator function whether the survival time is observed ($\delta_i = 1$) or censored ($\delta_i = 0$), $t_i$ is the survival time for the $i^{th}$ patient and $\Omega(t_i)$ is the risk set at time $t_i$, i.e., the set of patients who still survived before $t_i$. $\hat{Y}_i^{(1)}$ denote the predicted survival risk of the $i^{th}$ patient. The loss function of grade classification is negative log likelihood:

$$l^{(2)} = \sum_{i=1}^{n} \frac{-\hat{Y}_{i,y_i}^{(2)}}{n} \tag{6}$$

where $y_i$ is ground-truth histologic grade label for the $i^{th}$ patient. $\hat{Y}_{i,y_i}^{(2)}$ denote that for the $i^{th}$ patient, only the log-softmax value corresponding to the true label is used to calculate the loss. There are many training strategies for multi-task learning such as joint training and alternate training. The two tasks used in our framework are closely related task. Survival analysis is a regression task that outputs a real-value to represent the hazard of patients, while grade classification outputs the group of patients (i.e., one of three integers) to represent the hazard of patients. Compared with the grade classification, we treat survival analysis as a more fine-grained task. Therefore, alternate training is a more suitable method for multi-task learning of cancer prognosis than joint training. The process of alternate training is defined as:

$$loss = \begin{cases} l^{(1)} & \text{if } c \text{ is odd} \\ l^{(2)} & \text{if } c \text{ is even} \end{cases} \quad (7)$$

where $c$ denote the current number of iterations. The total number of iterations is determined by the batch size and the total number of training samples.

## 3. Results and Discussion

### 3.1. Evaluation metrics

The concordance index (C-index) is a generalization of the area under the ROC curve that can take into account censored data, which represents the model's ability to correctly provide a reliable ranking of the survival times based on the individual risk scores. It can be computed with the following formula:

$$\text{C-index} = \frac{\sum_{i,j} 1_{t_j < t_i} \cdot 1_{\hat{y}_j^{(1)} > \hat{y}_i^{(1)}} \cdot \delta_j}{\sum_{i,j} 1_{t_j < t_i} \cdot \delta_j} \quad (8)$$

where the term $1_{t_j < t_i}$ denotes that when $t_j < t_i$, $1_{t_j < t_i} = 1$, otherwise it is equal to 0. The implementation of C-index is from lifelines (https://lifelines.readthedocs.io/en/latest/).

In the glioma dataset, there are three types of labels for cancer grade classification. Therefore, we employ the micro-average to calculate the Area Under the Receiver Operating Characteristic Curve (micro-AUC), average precision (micro-AP), and F1 score (micro-F1) for performance evaluation. The micro-average aggregates the contributions of all classes to compute the average metric, which is suitable for class imbalanced data. The implementation of these metrics is from scikit-learn (http://scikit-learn.org).

### 3.2. Comparison of MultiCoFusion With Other Multi-modal Methods

We compare our framework MultiCoFusion with several traditional methods and two deep learning methods of multi-modal fusion. For traditional methods, it is not feasible to directly use high-dimensional data as input. So we first extract 150 image features and 30 genomic features [5]. The multi-modal feature extraction method has been used in many cancers such as KIRC, KIRP, and LUSC and demonstrates a strong ability in survival analysis tasks [5, 30]. Then we use two methods to perform feature fusion: concatenation and CCA (the number of components to keep is 10).

Finally we use LASSO-Cox model to predict survival outcomes and Multi-layer Perceptron (MLP), Logistic Regression (LR), and PWMK to predict cancer grade. MLP and LR are implemented using scikit-learn 0.24.0 with default parameters [27, 1]. PWMK [37] is a heuristic MKL algorithm that assigns the weights according to the performance of individual kernels. In this paper, we choose Gaussian kernel as base kernel. The image and genomic features are processed by ten different Gaussian kernels respectively. The implementation of PWMK is from Lauriola et al. [16]. For deep learning methods, Pathomic Fusion [4] and MultimodalPrognosis [3] are used to compare with our framework. They fuse multi-modal representations via Kronecker Product and Gating-Based Attention (KP&GBA) and mean respectively. All comparison algorithms are trained and tested using the same glioma dataset (Section.2.1). All deep learning models (i.e., MultimodalPrognosis, Pathomic Fusion and MultiCoFusion) are trained for 30 epochs using the Adam optimizer, dropout probability 0.25, and a linearly decaying learning rate scheduler. The learning rate is initialized to 0.0001, the weight decay is 0.0004, and the batch size is 32.

Experiment results are showed in Table.3. The feature extraction method proposed by Cheng et al. [5] provides strong support for survival analysis task (C-index=0.831±0.052), even if multi-modal features are fused by concatenating. But this type of feature is extracted for survival analysis, so it is not very effective in grade task. Based on these features, CCA further improves the performance of traditional methods in grade task. The performance of MultimodalPrognosis is worse than some traditional methods, we think this is because the architecture of MultimodalPrognosis is very simple. Compared with all traditional algorithms, Pathomic Fusion has achieved a good performance, which demonstrates that suitable deep learning method can learn better representations than traditional feature extraction method. Through more powerful image and mRNA feature representation learning methods (i.e., ResNet-152 and SGCN), and multi-task alternate learning scheme, our proposed framework MultiCoFusion has achieved the best performance on both tasks. In addition, we report the F1 score for Grade IV, because this is the most serious situation, so it is important to make an accurate prediction for it. We find that CCA helps all traditional methods achieve almost perfect results in this metric. Note that the F1 score of 1.000 does not mean that the model can predict perfectly. All of our experimental results keep third digit after the decimal point, so they are rounded to 1.000.

From the above comparison, it can be found that the performance of MultiCoFusion and Pathomic Fusion is better than all other models, so we futher report their accuracy and confusion matrix. We utilize the accuracy_score of sklearn-0.24.0 to calculate their accuracy. The accuracy of MultiCoFusion and Pathomic Fusion is 0.756 ± 0.032 and 0.716 ± 0.035, respectively. Our experiments are repeated 15 times, so we present the sum of confusion matrices for the 15 experiments in Table.1 and Table.2. We can find that the performance of MultiCoFusion is better than Pathomic Fusion

**Table 1**
The Confusion Matrix of MultiCoFusion

|      | Prediction | | |
|------|-----|-----|-----|
|      | 850 | 232 | 0   |
| True | 418 | 721 | 0   |
|      | 2   | 0   | 451 |

**Table 2**
The Confusion Matrix of Pathomic Fusion

|      | Prediction | | |
|------|-----|-----|-----|
|      | 763 | 319 | 0   |
| True | 443 | 693 | 3   |
|      | 1   | 0   | 452 |

in both metrics.

### 3.3. Comparison Between Different Situations

MultiCoFusion can be divided into different forms to adapt to different situations, including Single-modal and Single-task (SMST), Multi-modal and Single-task (MMST), and Single-modal and Multi-task (SMMT). We construct some experiments to compare the model performance between the subtype of MultiCoFusion and other models in different situations. In SMST, we examine the performance of different data modalities and different representation methods. In MMST, we examine the performance of different fusion methods and the combination of different representation methods. In SMMT, we examine the performance of different multi-task learning methods. Finally, we compare the optimal results in SMST, MMST, and SMMT with MultiCoFusion, which belongs to the situation of Multi-modal and Multi-task (MMMT).

#### 3.3.1. Single-modal and Single-task (SMST)

In MultiCoFusion, we use ResNet-152 and SGCN to learning the representation of histopathological images and mRNA expression data respectively. The VGG-19, ResNet-18, and Genomic SNN (GSNN) [4] are used to compare with our methods. The three imaging models (i.e., VGG-19, ResNet-18, and ResNet-152) are all pre-trained on ImageNet. They are followed by a four-layer FCNN with dropout to compress the representations into low-dimension (32 dimensions) for survival analysis or grade classification. GSNN consists of five layers FCNN, each layer is followed by Exponential Linear Unit (ELU) activation and Alpha Dropout to ensure the self-normalization property. For a fair comparison with SGCN, the input and output dimension of the first layer of GSNN are the same. For SGCN, the first layer is defined by Eq.(2), and then a four-layer FCNN, each layer is followed by Scaled Exponential Linear Unit (SELU) activation and Alpha Dropout. The final dimension of both GSNN and SGCN is the same as image models (i.e., 32 dimensions). The optimization process of all models is similar to Chen et al. [4]. Specifically, These models are trained for 50 epochs using the Adam optimizer, dropout probability 0.25, and a linearly decaying learning rate scheduler. For image models, the learning rate is initialized to 0.0005, the weight decay is 0.0004, and the batch size is 8. For genomic models, the learning rate is initialized to 0.002, the weight decay is 0.0005, and the batch size is 64.

The results are showed in Table.4. First of all, we observe that the mRNA expression data may provide more information than the histopathological images, because the performance of the mRNA expression models are better than the models of histopathological images in both tasks. This situation is different from Chen et al. [4]. In their paper, the genomic data achieves better performance for survival analysis, while histopathological images achieve better performance for grade. We think this is because they only use 320 genomic features in their experiments, while our experiments use more than 10,000 genomic features. To this end, we also test the GSNN using 320 features. The experimental results show that the C-index of GSNN (with 320 features) is better than VGG-19 (0.799 ± 0.015 v.s. 0.773 ± 0.017), but the micro-AUC is worse than VGG-19 (0.846 ± 0.013 v.s. 0.866 ± 0.012). This result is consistent with Chen et al. [4], and demonstrates that more mRNA features can improve the performance of both tasks. In the models of histopathological images, ResNet-152 is better at performing survival analysis task, while VGG-19 is better at performing grade task. In the models of mRNA expression data, the performance of SGCN is better than GSNN and has achieved the best results in both tasks. This result indicates that gene interaction information can indeed improve the representational ability of a model, thereby affecting the results of survival analysis and grade.

#### 3.3.2. Multi-modal and Single-task (MMST)

In the multi-modal fusion part of MultiCoFusion, we concatenate the representations of multiple modalities (1000 dimensions per modal) into a whole, and then a three-layer FCNN is employed to learn a fused representation (32 dimensions). The output dimension of VGG-19 is $512 \times 7 \times 7$, so an additional FCNN is used to compress it into 1000 dimensions. The output dimension of ResNet-152 is 1000 dimension, so its output can be used directly. For GSNN and SGCN, we first take the representations of their first hidden layer, and then use Eq.(3) (i.e., a FCNN) to compress them into 1000 dimensions. We compare our fusion method with Pathomic Fusion [4], which fuses multi-modal representations (32 dimensions) via Kronecker Product and Gating-Based Attention (KP&GBA). The architecture of representation model of KP&GBA are the same as SMST, which output 32-dimensional representations. There are two differences between KP&GBA and our concatenation. The first is the different ways of fusion, and the second is the different feature dimensions during fusion. In order to make a fairer comparison, we directly replace the KP&GBA with concatenation to form a new fusion method called Late Concatenation, which eliminating the second difference. For more clarity, we call our fusion method as Early Concatenation. All models are trained for 30 epochs using the Adam optimizer, dropout probability 0.25, and a linearly decaying learning rate scheduler. The learning rate is initialized to 0.0001, the weight decay is 0.0004, and the batch size is 32.

**Table 3**
Comparison of MultiCoFusion with Other Multi-modal Methods

| Model | Fusion Method | C-index | micro-AUC | micro-AP | micro-F1 | F1 (Grade IV) |
|---|---|---|---|---|---|---|
| LASSO-Cox[5] | Concatenation | 0.831±0.052 | - | - | - | - |
| MLP | Concatenation | - | 0.710±0.037 | 0.512±0.041 | 0.613±0.049 | 0.626±0.089 |
| LR | Concatenation | - | 0.710±0.037 | 0.513±0.042 | 0.614±0.049 | 0.649±0.102 |
| PWMK | Concatenation+MKL | - | 0.758±0.039 | 0.575±0.056 | 0.678±0.052 | 0.907±0.062 |
| LASSO-Cox | CCA | 0.823±0.046 | - | - | - | - |
| MLP | CCA | - | 0.784±0.037 | 0.610±0.055 | 0.712±0.050 | 1.000±0.000 |
| LR | CCA | - | 0.801±0.031 | 0.634±0.047 | 0.735±0.041 | 1.000±0.000 |
| PWMK | CCA+MKL | - | 0.799±0.035 | 0.633±0.052 | 0.733±0.046 | 1.000±0.000 |
| MultimodalPrognosis[3] | Mean | 0.731±0.023 | 0.732±0.031 | 0.552±0.042 | 0.526±0.037 | 0.309±0.126 |
| Pathomic Fusion[4] | KP&GBA | 0.832±0.018 | 0.852±0.023 | 0.699±0.038 | 0.716±0.035 | 0.996±0.005 |
| MultiCoFusion | Concatenation | **0.857**±0.015 | **0.923**±0.014 | **0.850**±0.027 | **0.759**±0.032 | 0.998±0.005 |

**Table 4**
Comparative Analysis of SMST Models

| Data Type | Representation Model | C-index | micro-AUC | micro-AP | micro-F1 | F1 (Grade IV) |
|---|---|---|---|---|---|---|
| Histopathological images | ResNet-18 | 0.769±0.016 | 0.852±0.013 | 0.725±0.023 | 0.656±0.020 | 0.810±0.030 |
| | ResNet-152 (ours) | 0.783±0.016 | 0.860±0.011 | 0.733±0.020 | 0.657±0.018 | 0.828±0.033 |
| | VGG-19 | 0.773±0.017 | 0.866±0.012 | 0.746±0.022 | 0.679±0.021 | 0.830±0.021 |
| mRNA expression | GSNN | 0.837±0.014 | 0.887±0.017 | 0.737±0.036 | 0.714±0.036 | 1.000±0.000 |
| | SGCN (ours) | **0.838**±0.018 | **0.897**±0.017 | **0.788**±0.035 | **0.730**±0.040 | 1.000±0.000 |

Experimental results are showed in Table.5. We can find that compare with SMST models, multi-modal fusion by KP&GBA (ResNet-152+GSNN) and Late Concatenation can improve the performance of survival analysis task in some cases (i.e.,

KP&GBA ResNet-152+GSNN and all of Late Concatenation). But the performance of all KP&GBA and Late Concatenation models is worse than single-modal models in grade task. Only the Early Concatenation can improve the performance in both tasks. This may imply that better grade results

**Table 5**
Comparative Analysis of MMST Models

| Fusion Method | Representation Model | C-index | micro-AUC | micro-AP | micro-F1 | F1 (Grade IV) |
|---|---|---|---|---|---|---|
| KP&GBA | VGG-19+GSNN | 0.837±0.018 | 0.830±0.023 | 0.702±0.038 | 0.703±0.030 | 0.884±0.026 |
| | VGG-19+SGCN | 0.834±0.017 | 0.862±0.024 | 0.755±0.040 | 0.712±0.040 | 0.922±0.028 |
| | ResNet-152+GSNN | 0.840±0.018 | 0.813±0.023 | 0.671±0.035 | 0.686±0.040 | 0.908±0.027 |
| | ResNet-152+SGCN | 0.836±0.019 | 0.852±0.029 | 0.730±0.048 | 0.714±0.039 | 0.916±0.027 |
| Late Concatenation | VGG-19+GSNN | 0.848±0.014 | 0.839±0.023 | 0.715±0.039 | 0.725±0.029 | 0.998±0.004 |
| | VGG-19+SGCN | 0.849±0.017 | 0.839±0.023 | 0.715±0.039 | 0.725±0.029 | 0.998±0.004 |
| | ResNet-152+GSNN | 0.846±0.010 | 0.826±0.022 | 0.712±0.035 | 0.723±0.028 | 0.966±0.014 |
| | ResNet-152+SGCN | **0.850**±0.018 | 0.850±0.022 | 0.723±0.044 | 0.724±0.038 | 0.910±0.049 |
| Early Concatenation | VGG-19+GSNN | 0.843±0.017 | 0.893±0.018 | 0.793±0.037 | 0.747±0.029 | 1.000±0.000 |
| | VGG-19+SGCN | **0.850**±0.016 | 0.915±0.013 | 0.839±0.025 | 0.750±0.032 | 0.997±0.007 |
| | ResNet-152+GSNN | 0.842±0.016 | 0.903±0.017 | 0.800±0.038 | **0.753**±0.028 | 0.999±0.002 |
| | ResNet-152+SGCN (ours) | **0.850**±0.017 | **0.918**±0.014 | **0.842**±0.028 | 0.743±0.034 | 0.999±0.002 |

**Table 6**
Comparative Analysis of SMMT Models

| Data Type | Representation Model | C-index | micro-AUC | micro-AP | micro-F1 | F1 (Grade IV) |
|---|---|---|---|---|---|---|
| Histopathological images | ResNet-18_AT | 0.773±0.018 | 0.859±0.012 | 0.738±0.023 | 0.659±0.023 | 0.828±0.030 |
| | ResNet-152_AT | 0.787±0.015 | 0.867±0.013 | 0.751±0.023 | 0.677±0.024 | 0.839±0.038 |
| | VGG-19_AT | 0.772±0.016 | 0.863±0.011 | 0.741±0.020 | 0.679±0.020 | 0.811±0.030 |
| mRNA expression | GSNN_AT | 0.835±0.013 | 0.901±0.015 | 0.806±0.033 | 0.716±0.035 | 1.000±0.000 |
| | SGCN_AT (ours) | **0.850**±0.013 | **0.915**±0.016 | **0.842**±0.027 | **0.753**±0.036 | 1.000±0.000 |
| | SGCN_ADD | 0.844±0.014 | 0.897±0.018 | 0.800±0.034 | 0.734±0.041 | 1.000±0.000 |
| | SGCN_HU | 0.845±0.014 | 0.894±0.018 | 0.784±0.032 | 0.736±0.038 | 1.000±0.000 |
| | SGCN_RHU | 0.845±0.014 | 0.894±0.018 | 0.784±0.032 | 0.736±0.038 | 1.000±0.000 |
| | SGCN_DWA | 0.845±0.014 | 0.907±0.017 | 0.822±0.030 | 0.740±0.038 | 1.000±0.000 |

**Table 7**
Comparison of MultiCoFusion with Other Situations

| Situation | Representation Model | C-index | micro-AUC | micro-AP | micro-F1 | F1 (Grade IV) |
|---|---|---|---|---|---|---|
| SMST | ResNet-152 | 0.783±0.016 | 0.860±0.011 | 0.733±0.020 | 0.657±0.018 | 0.828±0.033 |
| | SGCN | 0.838±0.018 | 0.897±0.017 | 0.788±0.035 | 0.730±0.040 | 1.000±0.000 |
| MMST | ResNet-152+SGCN | 0.850±0.017 | 0.918±0.014 | 0.842±0.028 | 0.743±0.034 | 0.999±0.002 |
| SMMT | ResNet-152_AT | 0.787±0.015 | 0.867±0.013 | 0.751±0.023 | 0.677±0.024 | 0.839±0.038 |
| | SGCN_AT | 0.850±0.013 | 0.915±0.016 | 0.842±0.027 | 0.753±0.036 | 1.000±0.000 |
| MMMT | MultiCoFusion (ResNet-152+SGCN AT) | **0.857**±0.015 | **0.923**±0.014 | **0.850**±0.027 | **0.759**±0.032 | 0.998±0.005 |

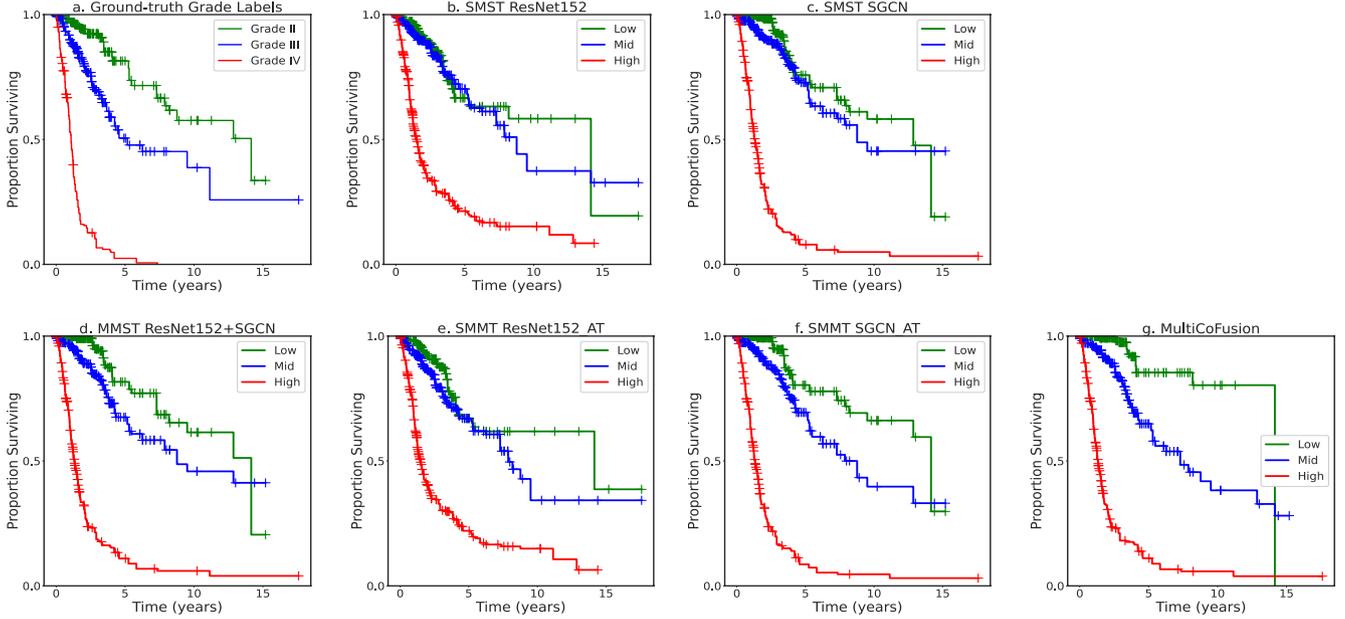

**Figure 2:** Kaplan-Meier (K-M) curves. a. Patients are divided into three groups according to the ground-truth grade labels and then perform Kaplan-Meier survival analysis. b–g. Patients are divided into three groups (i.e., Low, Mid, and High) according to the 33-66 percentile of hazard predictions. The proposed MultiCoFusion framework could easily separate the patients into low, mid and high risk groups

can be obtained when fusion is performed at a lower level (or a less abstract level). We also find that in survival analysis, the models with the best results all include the SGCN, which demonstrates the strong representational ability of SGCN.

### 3.3.3. Single-modal and Multi-task (SMMT)

For SMMT, we test all single-modal models of Section 3.3.1 in a multi-task situation. The difference is that a regressor and a classifier is connected to the 32-dimensional representation, and then the models are trained through alternate training (AT). Besides, we conduct four joint training experiments for SGCN, including simply adding task-specific cost functions together (ADD), Homoscedastic Uncertainty (HU) [6], Revised Homoscedastic Uncertainty (RHU) [18] and Dynamic Weight Average (DWA) [19]. All models are trained follow the same settings in SMST.

The results are showed in Table 6. Compared with the models in SMST, the performance of all models is improved with the help of multi-task learning for both tasks. For SGCN, the four joint training methods can improve performance, but the improvement is not as much as alternate training.

### 3.3.4. Comparison of MultiCoFusion with Other Situations

As far as we know, there is no model that considers both multi-modal and multi-task, so we compare our MultiCoFusion model with the optimal models in SMST, MMST, and SMMT, that is ResNet-152 and SGCN in SMST, ResNet-152+SGCN (Early Concatenation) in MMST, and ResNet-152_AT and SGCN_AT in SMMT.

Experimental results are shown in Table.7. As discussed above, compared with SMST models, both multi-modal and multi-task can improve performance. The optimal SMMT model SGCN_AT achieves a similar result to the optimal MMST model ResNet-152+SGCN. This may mean that even when there is no multi-modal data, we can enhance the performance of survival analysis and grade through multi-task learning. The performance of MultiCoFusion is better than all the optimal models, which shows the power of various strategies in our framework.

To further investigate the ability of MultiCoFusion for survival analysis, we plot Kaplan-Meier (K-M) curves of our model against the other models and ground-truth grade labels. The risk categories (low, mid, and high) are established by the 33-66 percentile of hazard predictions. It can be found from Figure.2.a that the two tasks, survival analysis and grade classification, have a very strong correlation. This is the cornerstone of our multi-task learning. From Figure.2.b-f, we observe that the K-M curves of low and mid risk group are intertwined, which indicates that the stratification of lower risk patients is difficult. While the K-M cures of high risk group are clearly distinguished. This phenomenon is consistent with other glioma-related papers [21, 4]. MultiCoFusion can alleviate the intertwined problem and clearly distinguish low, mid and high risk groups. It is worth noting that the Low curve of Figure.2.f drop to zero, because a death happens after the last censoring. For cancer grade classification, Grade II and Grade III are more difficult to distinguish than Grade IV. We plot Receiver Operating Characteristic Curves (ROCs) of Grade II and Grade III in Figure.3. The confidence interval is representative of the 15 times repetitions. Our method has greater AUCs in all cases.

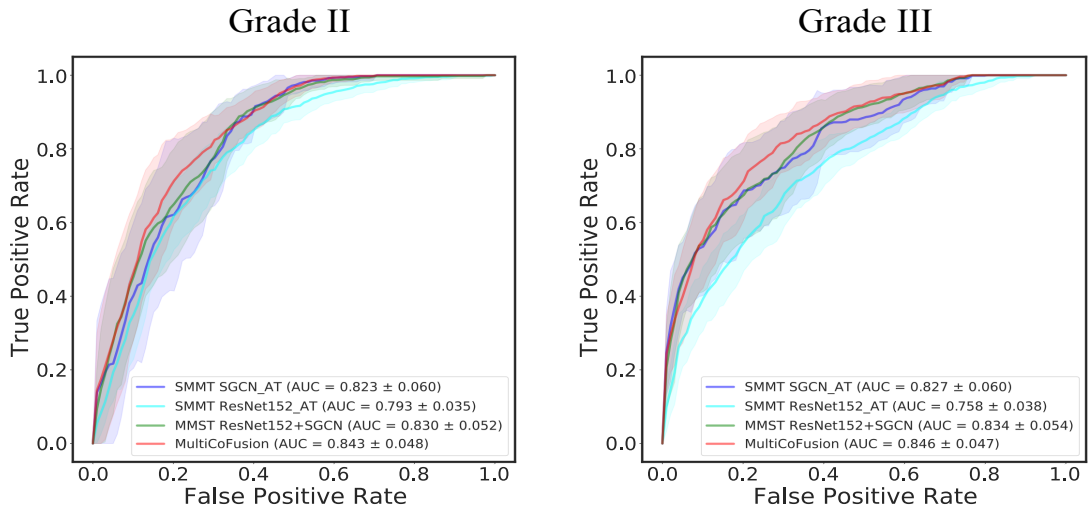

**Figure 3:** The ROCs and AUC scores for Grade II and Grade III. The confidence interval is representative of the 15 times repetitions. The performance of MultiCoFusion is better than other methods.

## 4. Conclusion

A large number of studies have shown that the fusion of multiple modalities can effectively improve performance of cancer diagnosis and prognosis prediction. However, they do not consider the correlation between different tasks. In this paper, we propose a multi-modal and multi-task deep learning framework (MultiCoFusion) to combine the power of different modalities and different tasks. A good representation is the basis of fusion and multi-task learning. We use ResNet-152 to extract the representation of histopathological images, and use SGCN to extract the representation of mRNA expression data. Compared with VGG-19, ResNet-152 has fewer parameters, lower complexity, but more layers. Compared with GSNN and FCNN, SGCN can model the interaction between features (i.e., genes). These representations are first concatenated, and then fused by a 3-layer FCNN, which is also the multi-task shared architecture. An alternate training strategy is used for training. Experiments on the glioma dataset demonstrate the strong representation ability of SGCN, the better performance of MultiCoFusion for both survival analysis and grade classification. Our framework may provide a new perspective for the analysis of the pathogenesis of cancer and has the potential to expand to more cancers and tasks. In addition, the trade-offs between the single-task learning and multi-task learning are an important research direction, and researchers can conduct in-depth studies on this.

## References

[1] Buitinck, L., Louppe, G., Blondel, M., Pedregosa, F., Mueller, A., Grisel, O., Niculae, V., Prettenhofer, P., Gramfort, A., Grobler, J., Layton, R., VanderPlas, J., Joly, A., Holt, B., Varoquaux, G., 2013. API design for machine learning software: experiences from the scikit-learn project, in: ECML PKDD Workshop: Languages for Data Mining and Machine Learning, pp. 108–122.

[2] Cerami, E., Gao, J., Dogrusoz, U., Gross, B.E., Sumer, S.O., Aksoy, B.A., Jacobsen, A., Byrne, C.J., Heuer, M.L., Larsson, E., Antipin, Y., Reva, B., Goldberg, A.P., Sander, C., Schultz, N., 2012. The cBio Cancer Genomics Portal: An Open Platform for Exploring Multidimensional Cancer Genomics Data. Cancer Discovery 2, 401–404. doi:10.1158/2159-8290.CD-12-0095.

[3] Cheerla, A., Gevaert, O., 2019. Deep learning with multimodal representation for pancancer prognosis prediction. Bioinformatics 35, i446–i454. doi:10.1093/bioinformatics/btz342.

[4] Chen, R.J., Lu, M.Y., Wang, J., Williamson, D.F.K., Rodig, S.J., Lindeman, N.I., Mahmood, F., 2020. Pathomic Fusion: An Integrated Framework for Fusing Histopathology and Genomic Features for Cancer Diagnosis and Prognosis. arXiv:1912.08937 [cs, q-bio] ArXiv: 1912.08937.

[5] Cheng, J., Zhang, J., Han, Y., Wang, X., Ye, X., Meng, Y., Parwani, A., Han, Z., Feng, Q., Huang, K., 2017. Integrative Analysis of Histopathological Images and Genomic Data Predicts Clear Cell Renal Cell Carcinoma Prognosis. Cancer Research 77, e91–e100. doi:10.1158/0008-5472.CAN-17-0313.

[6] Cipolla, R., Gal, Y., Kendall, A., 2018. Multi-task Learning Us-

ing Uncertainty to Weigh Losses for Scene Geometry and Semantics, in: 2018 IEEE/CVF Conference on Computer Vision and Pattern Recognition, IEEE, Salt Lake City, UT, USA. pp. 7482–7491. doi:10.1109/CVPR.2018.00781.

[7] Das, J., Yu, H., 2012. Hint: High-quality protein interactomes and their applications in understanding human disease. BMC systems biology 6, 92.

[8] Deng, J., Zeng, W., Kong, W., Shi, Y., Mou, X., Guo, J., 2019. Multi-Constrained Joint Non-Negative Matrix Factorization with Application to Imaging Genomic Study of Lung Metastasis in Soft Tissue Sarcomas. IEEE Transactions on Biomedical Engineering , 1–1.

[9] Deng, J., Zeng, W., Luo, S., Kong, W., Shi, Y., Li, Y., Zhang, H., 2021. Integrating multiple genomic imaging data for the study of lung metastasis in sarcomas using multi-dimensional constrained joint non-negative matrix factorization. Information Sciences 576, 24–36.

[10] Du, L., Liu, K., Zhu, L., Yao, X., Risacher, S.L., Guo, L., Saykin, A.J., Shen, L., Initiative, A.D.N., 2019. Identifying progressive imaging genetic patterns via multi-task sparse canonical correlation analysis: a longitudinal study of the ADNI cohort. Bioinformatics 35, i474–i483. doi:10.1093/bioinformatics/btz320.

[11] Gao, J., Aksoy, B.A., Dogrusoz, U., Dresdner, G., Gross, B., Sumer, S.O., Sun, Y., Jacobsen, A., Sinha, R., Larsson, E., Cerami, E., Sander, C., Schultz, N., 2013. Integrative Analysis of Complex Cancer Genomics and Clinical Profiles Using the cBioPortal. Science Signaling 6, pl1–pl1. doi:10.1126/scisignal.2004088.

[12] Hao, J., Kosaraju, S.C., Tsaku, N.Z., Song, D.H., Kang, M., 2019. PAGE-Net: Interpretable and Integrative Deep Learning for Survival Analysis Using Histopathological Images and Genomic Data, in: Biocomputing 2020, WORLD SCIENTIFIC, Kohala Coast, Hawaii, USA. pp. 355–366. doi:10.1142/9789811215636_0032.

[13] He, K., Zhang, X., Ren, S., Sun, J., 2015. Deep Residual Learning for Image Recognition. arXiv:1512.03385 [cs] ArXiv: 1512.03385.

[14] Klambauer, G., Unterthiner, T., Mayr, A., Hochreiter, S., 2017. Self-Normalizing Neural Networks , 10.

[15] Kong, Y., Yu, T., 2018. A graph-embedded deep feedforward network for disease outcome classification and feature selection using gene expression data. Bioinformatics 34, 3727–3737. doi:10.1093/bioinformatics/bty429.

[16] Lauriola, I., Aiolli, F., 2020. Mklpy: a python-based framework for multiple kernel learning. arXiv preprint arXiv:2007.09982 .

[17] Lee, M., Han, S.W., Seok, J., 2019. Prediction of survival risks with adjusted gene expression through risk-gene networks , 9.

[18] Liebel, L., Körner, M., 2018. Auxiliary Tasks in Multi-task Learning. arXiv:1805.06334 [cs] ArXiv: 1805.06334.

[19] Liu, S., Johns, E., Davison, A.J., 2019. End-To-End Multi-Task Learning With Attention, in: 2019 IEEE/CVF Conference on Computer Vision and Pattern Recognition (CVPR), IEEE, Long Beach, CA, USA. pp. 1871–1880. doi:10.1109/CVPR.2019.00197.

[20] Louis, D.N., Perry, A., Reifenberger, G., von Deimling, A., Figarella-Branger, D., Cavenee, W.K., Ohgaki, H., Wiestler, O.D., Kleihues, P., Ellison, D.W., 2016. The 2016 World Health Organization Classification of Tumors of the Central Nervous System: a summary. Acta Neuropathologica 131, 803–820.

[21] Mobadersany, P., Yousefi, S., Amgad, M., Gutman, D.A., Barnholtz-Sloan, J.S., Velázquez Vega, J.E., Brat, D.J., Cooper, L.A.D., 2018. Predicting cancer outcomes from histology and genomics using convolutional networks. Proceedings of the National Academy of Sciences 115, E2970–E2979. doi:10.1073/pnas.1717139115.

[22] Niazi, M.K.K., Yao, K., Zynger, D.L., Clinton, S.K., Chen, J., Koyuturk, M., LaFramboise, T., Gurcan, M., 2017. Visually Meaningful Histopathological Features for Automatic Grading of Prostate Cancer. IEEE Journal of Biomedical and Health Informatics 21, 1027–1038. doi:10.1109/JBHI.2016.2565515.

[23] Ning, Z., Pan, W., Chen, Y., Xiao, Q., Zhang, X., Luo, J., Wang, J., Zhang, Y., 2020. Integrative analysis of cross-modal features for the prognosis prediction of clear cell renal cell carcinoma. Bioinformatics 36, 2888–2895. doi:10.1093/bioinformatics/btaa056.

[24] Ning, Z., Xiao, Q., Feng, Q., Chen, W., Zhang, Y., 2021. Relation-induced multi-modal shared representation learning for alzheimer's disease diagnosis. IEEE Transactions on Medical Imaging 40, 1632–1645.

[25] Ning, Z., Zhang, X., Tu, C., Feng, Q., Zhang, Y., 2019. Multiscale Context-Cascaded Ensemble Framework (MsC² EF): Application to Breast Histopathological Image. IEEE Access 7, 150910–150923. doi:10.1109/ACCESS.2019.2946478.

[26] Nir, G., 2018. Automatic grading of prostate cancer in digitized histopathology images: Learning from multiple experts. Medical Image Analysis , 14.

[27] Pedregosa, F., Varoquaux, G., Gramfort, A., Michel, V., Thirion, B., Grisel, O., Blondel, M., Prettenhofer, P., Weiss, R., Dubourg, V., Vanderplas, J., Passos, A., Cournapeau, D., Brucher, M., Perrot, M., Duchesnay, E., 2011. Scikit-learn: Machine learning in Python. Journal of Machine Learning Research 12, 2825–2830.

[28] Qi, Y., Tastan, O., Carbonell, J.G., Klein-Seetharaman, J., Weston, J., 2010. Semi-supervised multi-task learning for predicting interactions between HIV-1 and human proteins. Bioinformatics 26, i645–i652. doi:10.1093/bioinformatics/btq394.

[29] Russakovsky, O., Deng, J., Su, H., Krause, J., Satheesh, S., Ma, S., Huang, Z., Karpathy, A., Khosla, A., Bernstein, M., Berg, A.C., Fei-Fei, L., 2015. ImageNet Large Scale Visual Recognition Challenge. arXiv:1409.0575 [cs] ArXiv: 1409.0575.

[30] Shao, W., Huang, K., Han, Z., Cheng, J., Cheng, L., Wang, T., Sun, L., Lu, Z., Zhang, J., Zhang, D., 2020. Integrative Analysis of Pathological Images and Multi-Dimensional Genomic Data for Early-Stage Cancer Prognosis. IEEE Transactions on Medical Imaging 39, 99–110. doi:10.1109/TMI.2019.2920608.

[31] Shen, L., Thompson, P.M., 2020. Brain Imaging Genomics: Integrated Analysis and Machine Learning. Proceedings of the IEEE 108, 125–162. doi:10.1109/JPROC.2019.2947272.

[32] Srinidhi, C.L., Ciga, O., Martel, A.L., 2021. Deep neural network models for computational histopathology: A survey. Medical Image Analysis 67, 101813. doi:https://doi.org/10.1016/j.media.2020.101813.

[33] Sun, D., Li, A., Tang, B., Wang, M., 2018. Integrating genomic data and pathological images to effectively predict breast cancer clinical outcome. Computer Methods and Programs in Biomedicine 161, 45–53. doi:10.1016/j.cmpb.2018.04.008.

[34] Sun, X., Panda, R., Feris, R., Saenko, K., 2020. Adashare: Learning what to share for efficient deep multi-task learning. Advances in Neural Information Processing Systems 33.

[35] Tabibu, S., Vinod, P.K., Jawahar, C.V., 2019. Pan-Renal Cell Carcinoma classification and survival prediction from histopathology images using deep learning. Scientific Reports 9, 10509. doi:10.1038/s41598-019-46718-3.

[36] Tan, K., Huang, W., Liu, X., Hu, J., Dong, S., 2021. A hierarchical graph convolution network for representation learning of gene expression data. IEEE Journal of Biomedical and Health Informatics , 1–1doi:10.1109/JBHI.2021.3052008.

[37] Tanabe, H., Bao Ho, T., Nguyen, C.H., Kawasaki, S., 2008. Simple but effective methods for combining kernels in computational biology, in: 2008 IEEE International Conference on Research, Innovation and Vision for the Future in Computing and Communication Technologies, pp. 71–78. doi:10.1109/RIVF.2008.4586335.

[38] Tepeli, Y.I., Ünal, A.B., Akdemir, F.M., Tastan, O., 2020. PAMOGK: a pathway graph kernel-based multiomics approach for patient clustering. Bioinformatics , btaa655doi:10.1093/bioinformatics/btaa655.

[39] Venugopalan, J., Tong, L., Hassanzadeh, H.R., Wang, M.D., 2021. Multimodal deep learning models for early detection of alzheimer's disease stage. Scientific reports 11, 1–13.

[40] Wan, T., Cao, J., Chen, J., Qin, Z., 2017. Automated grading of breast cancer histopathology using cascaded ensemble with combination of multi-level image features. Neurocomputing 229, 34–44. doi:10.1016/j.neucom.2016.05.084.

[41] Wang, B., Mezlini, A.M., Demir, F., Fiume, M., Tu, Z., Brudno, M., Haibe-Kains, B., Goldenberg, A., 2014. Similarity network fusion

for aggregating data types on a genomic scale. Nature Methods 11, 333–337. doi:10.1038/nmeth.2810.

[42] Wang, X., Zhang, Y., Ren, X., Zhang, Y., Zitnik, M., Shang, J., Langlotz, C., Han, J., 2018. Cross-type biomedical named entity recognition with deep multi-task learning. Bioinformatics 35, 1745–1752. doi:10.1093/bioinformatics/bty869.

[43] Wulczyn, E., Steiner, D.F., Xu, Z., Sadhwani, A., Wang, H., Flament-Auvigne, I., Mermel, C.H., Chen, P.H.C., Liu, Y., Stumpe, M.C., 2020. Deep learning-based survival prediction for multiple cancer types using histopathology images. PLOS ONE , 18.

[44] Xie, L., He, S., Zhang, Z., Lin, K., Bo, X., Yang, S., Feng, B., Wan, K., Yang, K., Yang, J., Ding, Y., 2020. Domain-adversarial multi-task framework for novel therapeutic property prediction of compounds. Bioinformatics 36, 2848–2855. doi:10.1093/bioinformatics/btaa063.

[45] Yasrebi, H., 2011. SurvJamda: an R package to predict patients' survival and risk assessment using joint analysis of microarray gene expression data. Bioinformatics 27, 1168–1169. doi:10.1093/bioinformatics/btr103.

[46] Yu, T., Kumar, S., Gupta, A., Levine, S., Hausman, K., Finn, C., 2020. Gradient surgery for multi-task learning. Advances in Neural Information Processing Systems .

[47] Zhang, Y., Li, A., He, J., Wang, M., 2020. A Novel MKL Method for GBM Prognosis Prediction by Integrating Histopathological Image and Multi-Omics Data. IEEE Journal of Biomedical and Health Informatics 24, 171–179. doi:10.1109/JBHI.2019.2898471.